\documentclass[runningheads,a4paper]{llncs}

\usepackage{amsmath}
\usepackage{amssymb}
\usepackage{graphicx}
\usepackage{color}
\usepackage{multirow}
\usepackage[colorinlistoftodos]{todonotes}
\usepackage{subcaption}
\usepackage{makecell}

\usepackage{url}
\urldef{\mailsa}\path|{AnonymousName}@AnonymousAddress|

\begin{document}

\mainmatter  

\title{A Lifelong Learning Approach to Brain MR Segmentation Across Scanners and Protocols}

\titlerunning{Lifelong Learning for MRI Segmentation Across Scanners and Protocols}

%
%
\author{Neerav Karani, Krishna Chaitanya, Christian Baumgartner, Ender Konukoglu}
%
\authorrunning{Lifelong Learning for MRI Segmentation Across Scanners and Protocols}

\institute{Computer Vision Lab, ETH Zurich, Zurich, Switzerland
}
%
%

\maketitle

\begin{abstract}
Convolutional neural networks (CNNs) have shown promising results on several segmentation tasks in magnetic resonance (MR) images.
However, the accuracy of CNNs may degrade severely when segmenting images acquired with different scanners and/or protocols as compared to the training data, thus limiting their practical utility.
We address this shortcoming in a lifelong multi-domain learning setting by treating images acquired with different scanners or protocols as samples from different, but related domains.
Our solution is a single CNN with shared convolutional filters and domain-specific batch normalization layers, which can be tuned to new domains with only a few ($\approx$ 4) labelled images.
Importantly, this is achieved while retaining performance on the older domains whose training data may no longer be available.
We evaluate the method for brain structure segmentation in MR images.
Results demonstrate that the proposed method largely closes the gap to the benchmark, which is training a dedicated CNN for each scanner.
\end{abstract}

\section{Introduction}
\begin{figure}
\setlength{\belowcaptionskip}{-20pt}
\centering
\includegraphics[trim = 40mm 0mm 40mm 10mm, clip, width=0.235\textwidth]{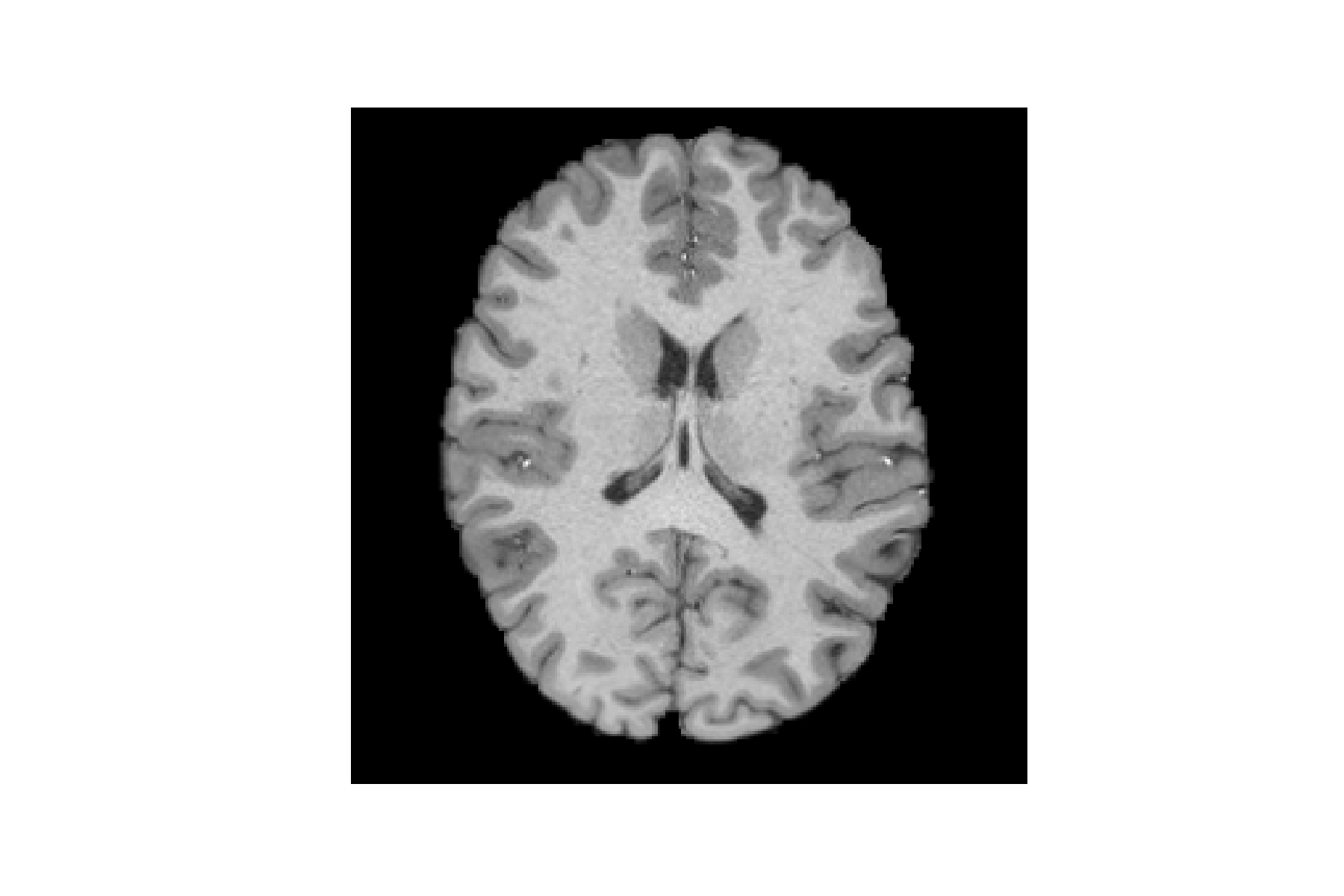}
\includegraphics[trim = 40mm 0mm 40mm 10mm, clip, width=0.235\textwidth]{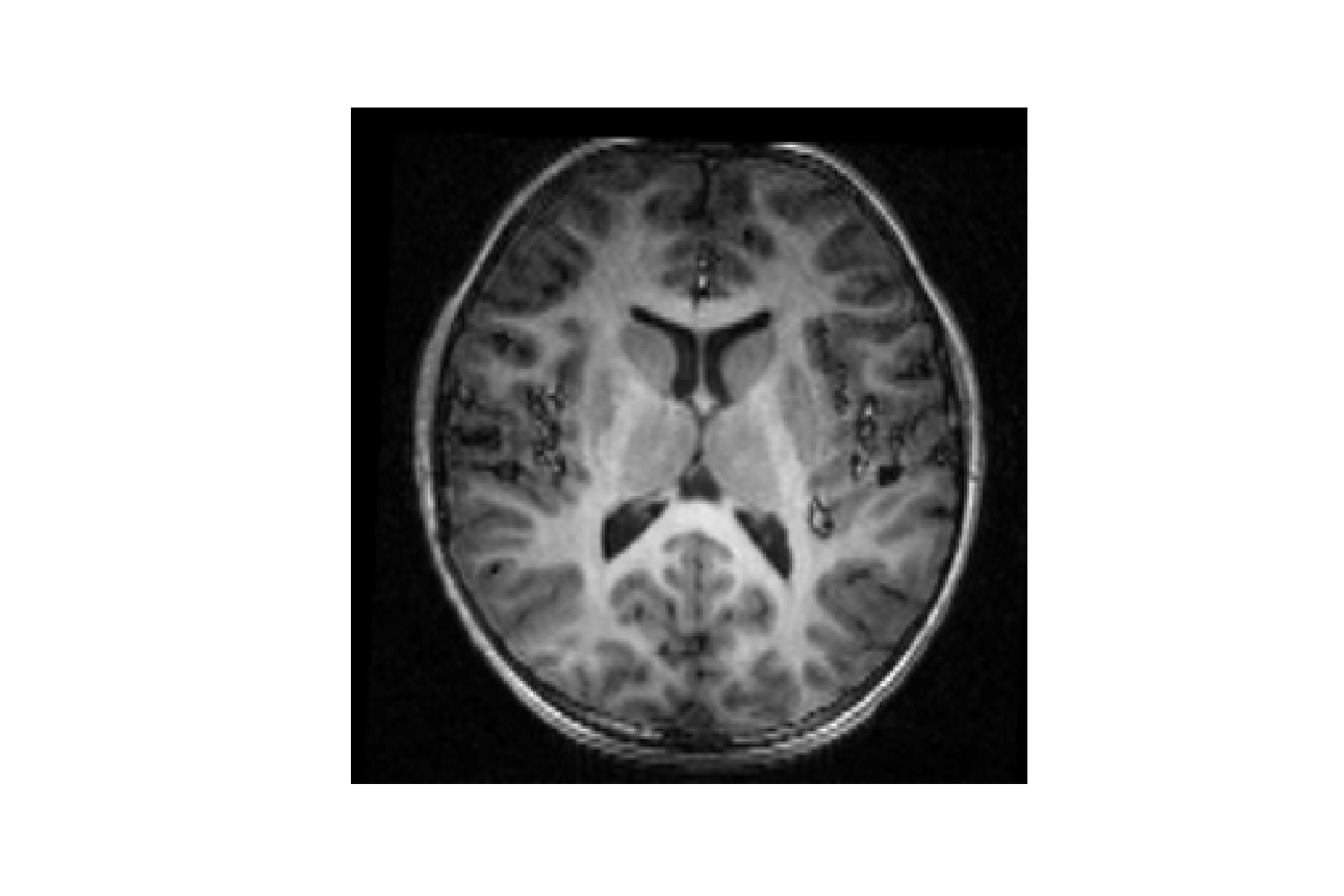}
\includegraphics[trim = 40mm 0mm 40mm 10mm, clip, width=0.235\textwidth]{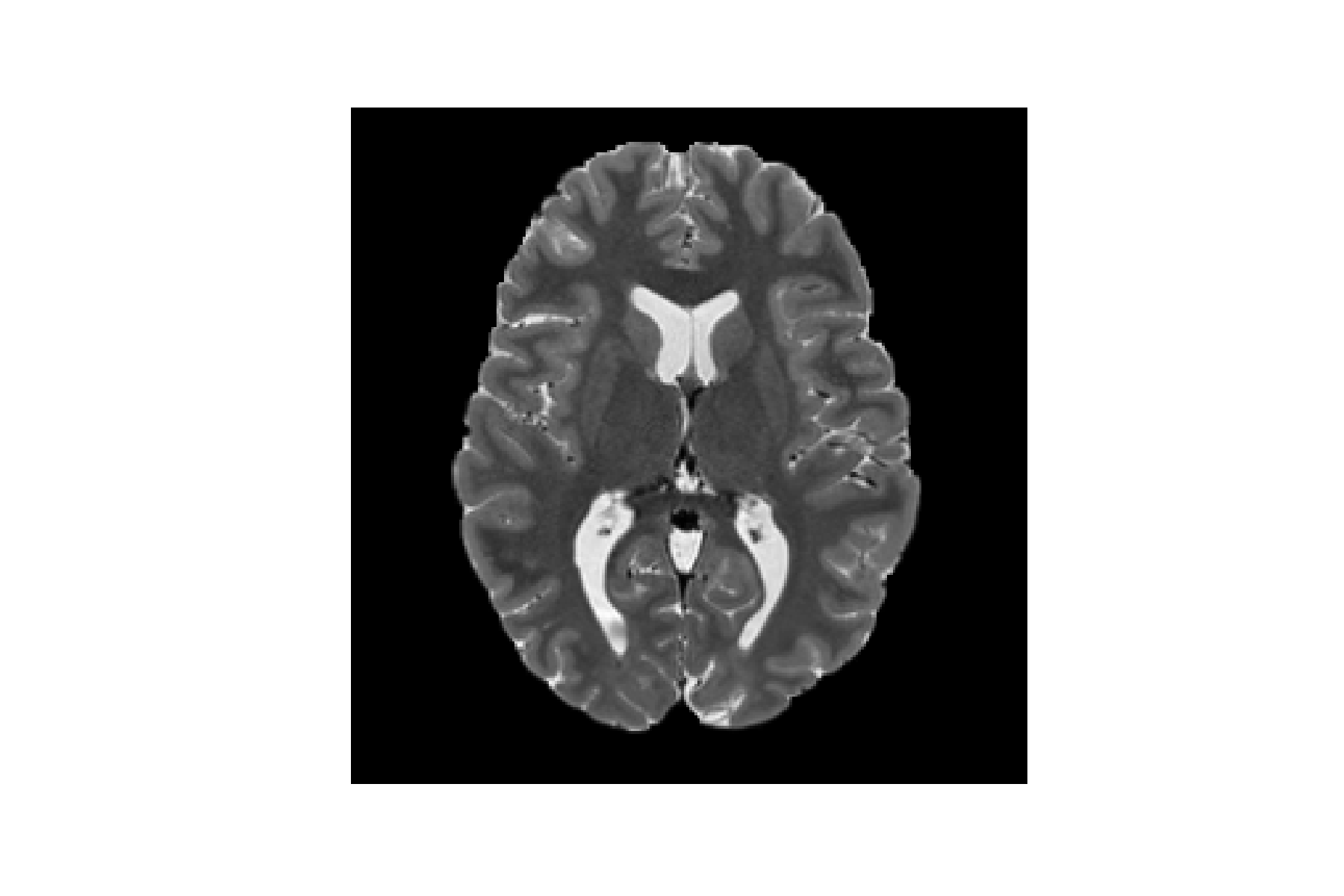}
\includegraphics[trim = 40mm 0mm 40mm 10mm, clip, width=0.235\textwidth]{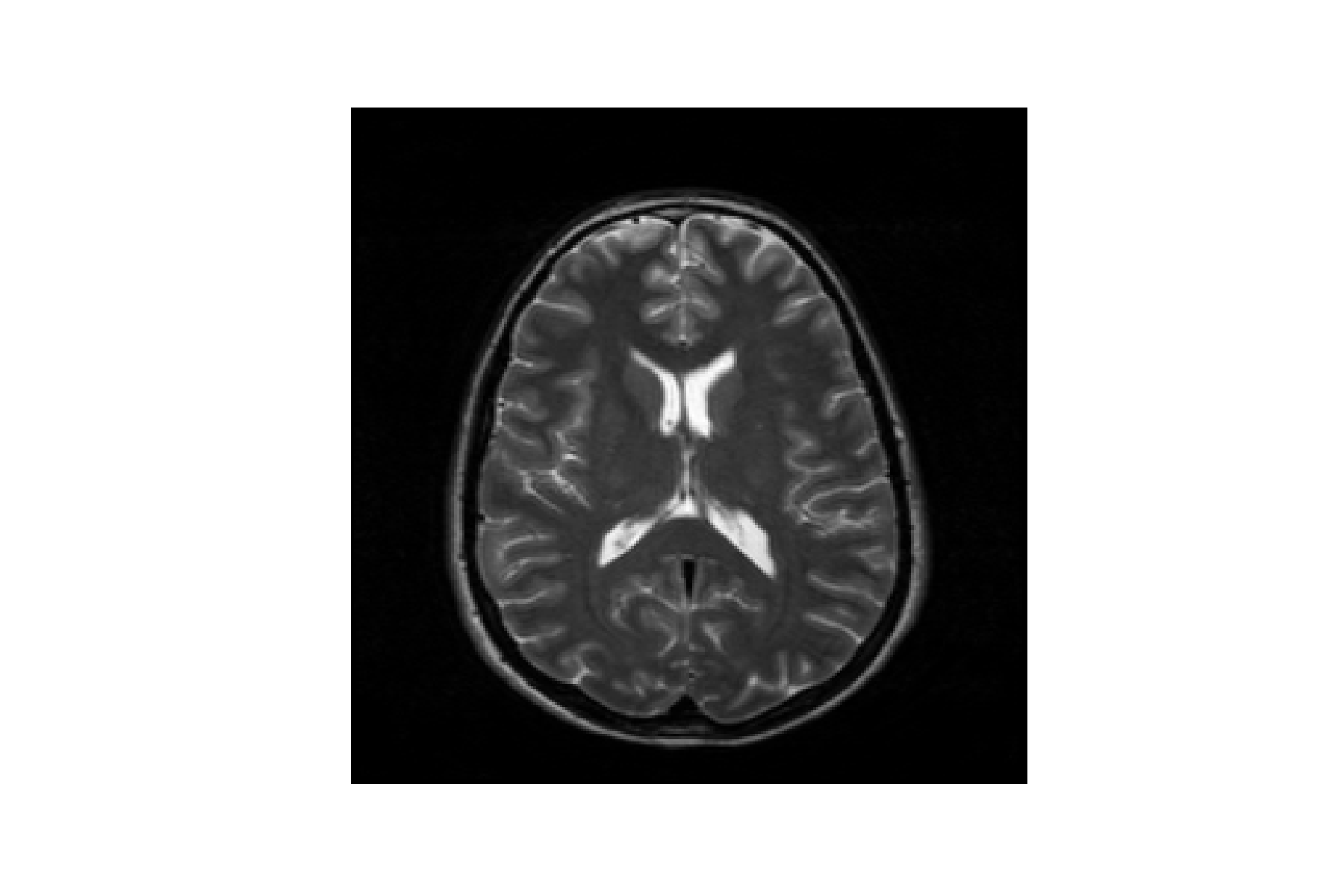}
\includegraphics[trim = 17mm 11mm 10mm 12mm, clip, width=0.235\textwidth]{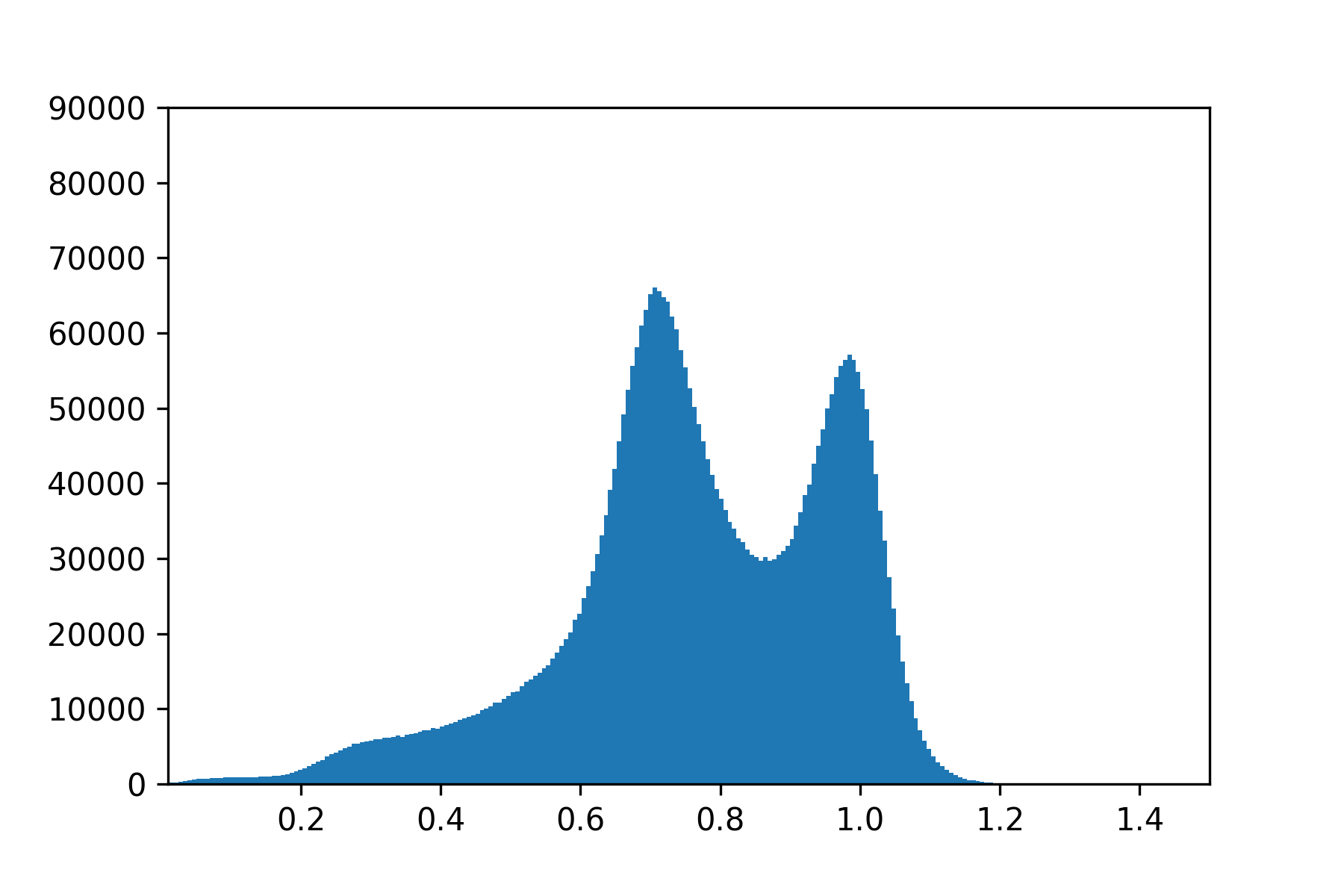}
\includegraphics[trim = 17mm 11mm 10mm 12mm, clip, width=0.235\textwidth]{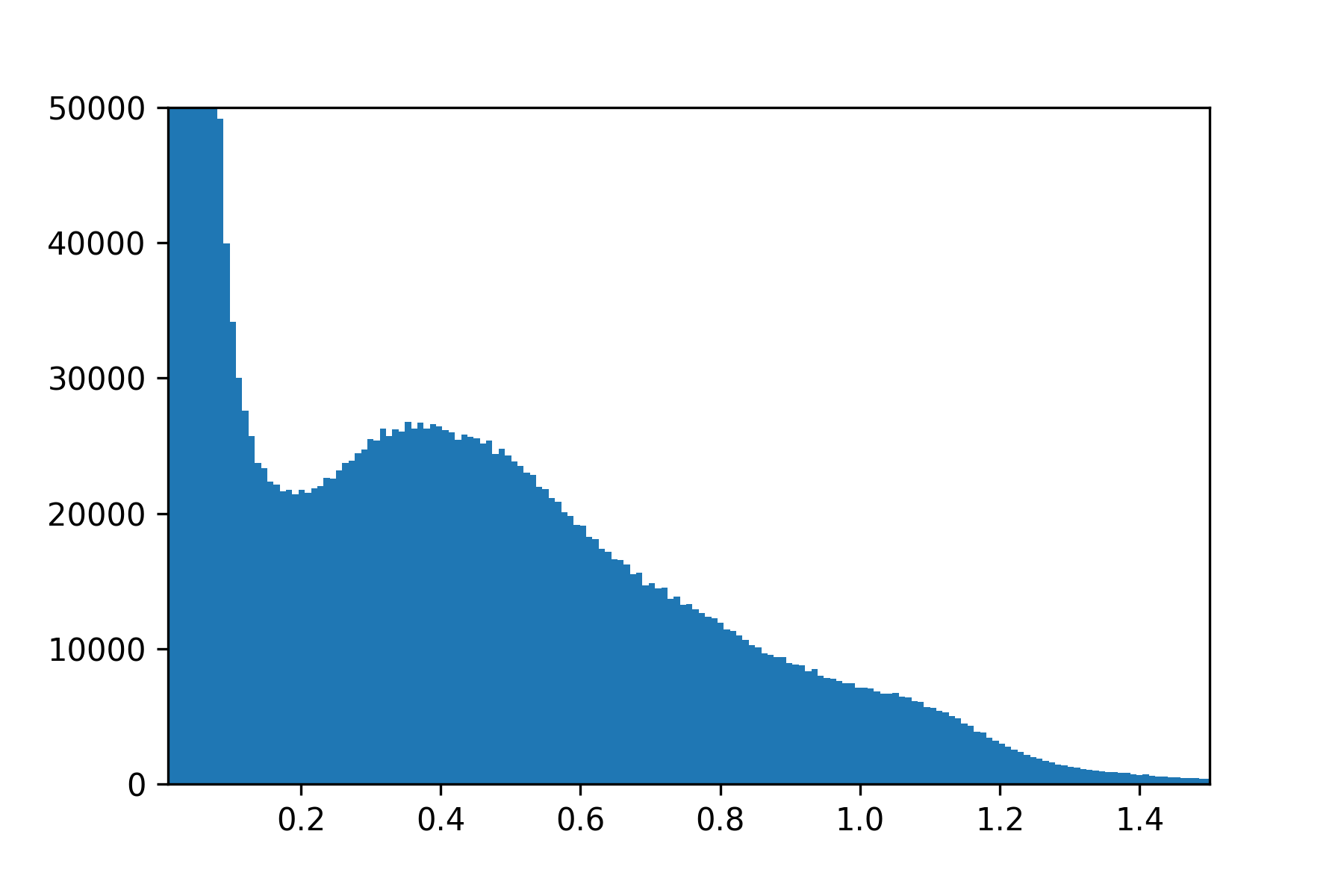}
\includegraphics[trim = 17mm 11mm 10mm 12mm, clip, width=0.235\textwidth]{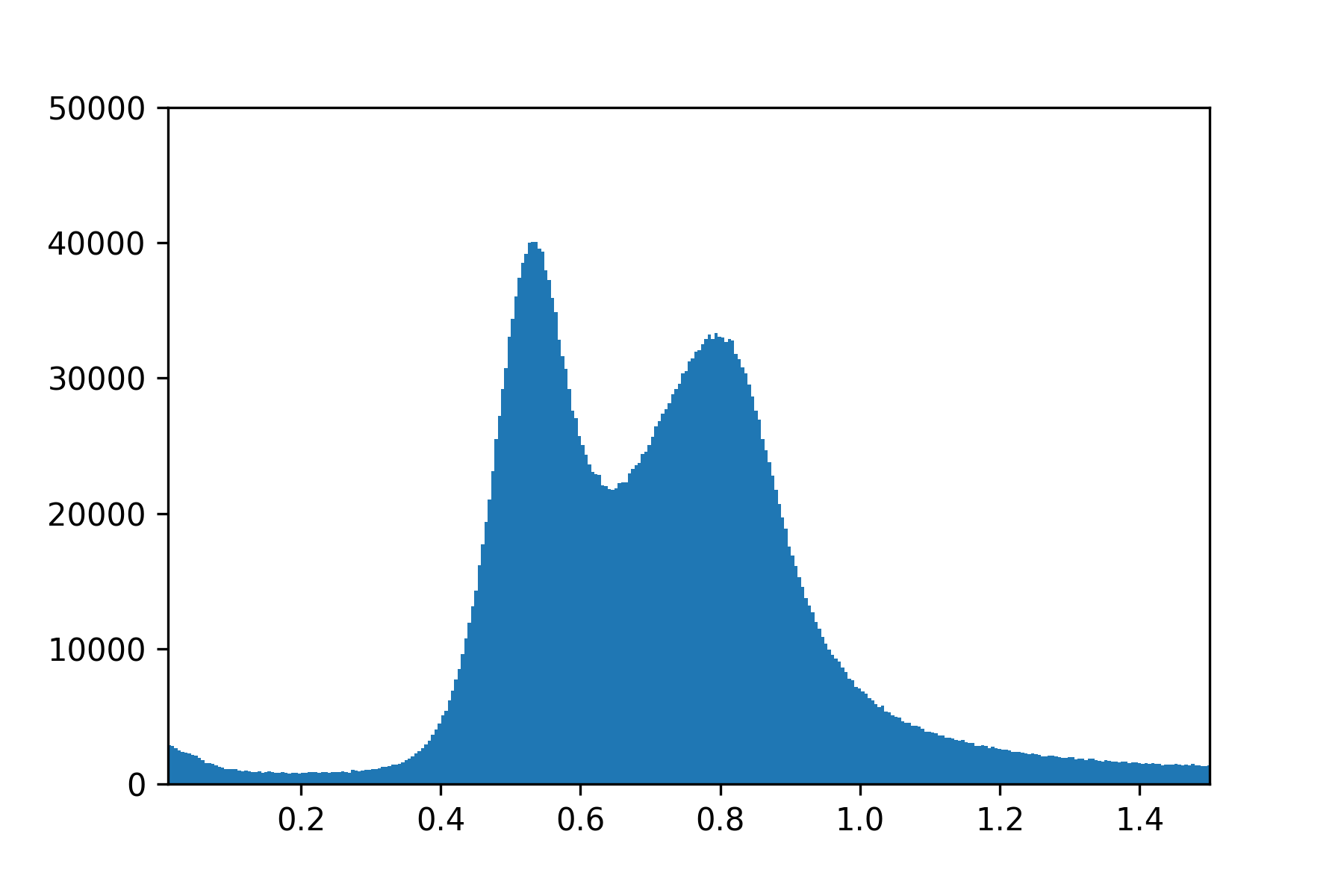}
\includegraphics[trim = 17mm 11mm 10mm 12mm, clip, width=0.235\textwidth]{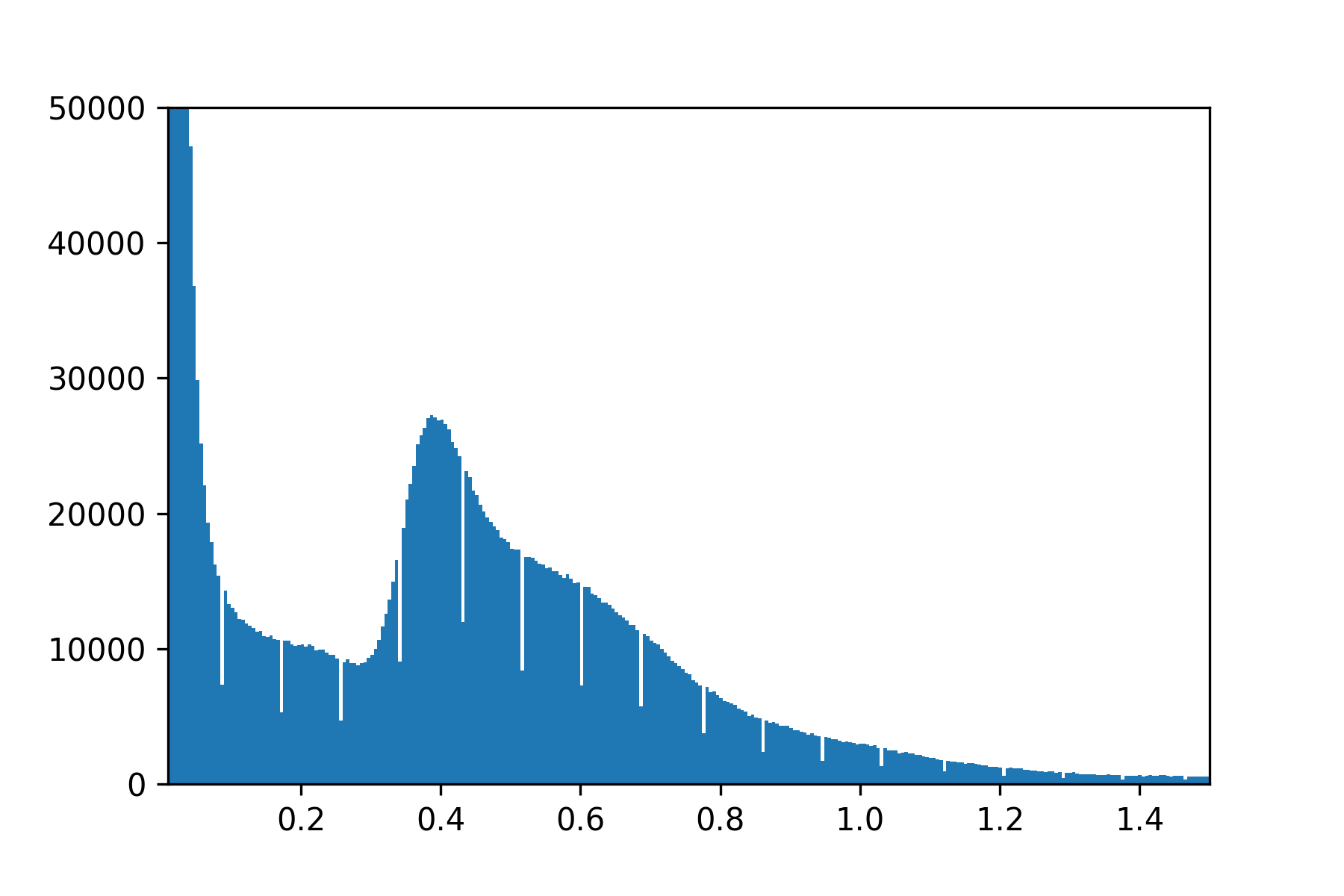}
\\ (a) \hspace{2.3cm} (b) \hspace{2.3cm} (c) \hspace{2.3cm} (d)
\caption{Image slices (top) and corresponding histograms (bottom) of normalized T1w (a,b) and T2w (c,d) MRIs from different scanners. Despite high-level information similarity, there exists considerable intensity and contrast differences, which segmentation algorithms are often sensitive to.}
\label{fig:domain_histograms}
\end{figure}

Segmentation of brain MR images is a critical step in many diagnostic and surgical applications.
Accordingly, several approaches have been proposed for tackling this problem such as atlas-based segmentation~\cite{fischl2012freesurfer}, methods based on machine learning techniques such as CNNs~\cite{ronneberger2015u}, among many others as detailed in this recent survey~\cite{despotovic2015mri}.
One of the important challenges in many MRI analysis tasks, including segmentation, is robustness to  differences in statistical characteristics of image intensities.
These differences might arise due to using different scanners in which factors like drift in scanner SNR over time~\cite{preboske2006common}, gradient non-linearities~\cite{jovicich2006reliability} and others play an important role. 
Intensity variations may even arise when scanning protocol parameters (flip angle, echo or repetition time, etc.) are slightly changed on the same scanner.
Fig.~\ref{fig:domain_histograms}(a,b) shows 2D slices from two T1-weighted MRI datasets from different scanners, along with their intensity histograms which show the aforementioned variations.
Segmentation algorithms are often very sensitive to such changes.
Furthermore, images acquired with different MR modalities, such as T1 and T2-weighted images, may have considerably high-level of similarity in image content (see Fig.~\ref{fig:domain_histograms}). 
While analyzing these images, humans can leverage such commonalities easily and it would be highly desirable if learning-based algorithms could mimic this trait.

In the parlance of transfer machine learning, images acquired from different scanners, protocols or similar MR modalities may be viewed as data points sampled from different domains, with the degree of domain shift potentially indicated by the differences in their intensity statistics.
This perspective motivates us to employ ideas from the literature of domain adaptation~\cite{pan2010survey}, multi-domain learning~\cite{dredze2010multi} and lifelong learning~\cite{thrun1998lifelong} to the problem of brain segmentation across scanners / protocols.
Domain adaptation / transfer learning refers to a situation where a learner trained on a source domain is able to perform well on a target domain, of which only a few labelled examples are available.
However, in this case, the performance on the source domain may not be necessarily maintained after adaptation.
Multi-domain learning aims to train a learner that can simultaneously perform well on multiple domains.
Finally, in lifelong learning, a multi-domain learner is able to incorporate new domains with only few labelled examples, while preserving performance on previous domains.

Variants of image intensity standardization~\cite{zhuge2009intensity,weisenfeld2004normalization} and atlas intensity renormalization~\cite{han2007atlas} have been proposed as pre-processing steps to insure conventional segmentation methods from inter-scanner differences.
Among learning methods based on hand-crafted features, transfer learning approaches have been employed for multi-site segmentation~\cite{van2015transfer} and classification~\cite{cheplygina2017transfer}.
While adaptive support vector machines used by~\cite{van2015transfer} may be adapted for new scanners in a lifelong learning sense, they are likely to be limited by the quality of the hand-crafted features.
Using CNNs,~\cite{kamnitsas2017unsupervised} propose to deal with inter-protocol differences by learning domain invariant representations.
This approach may be limited to work with the least common denominator between the domains, while, as shown in~\cite{bilen2017universal}, providing a few separate parameters for each domain allows for learning of domain specific nuances.
Further, it is unclear how~\cite{kamnitsas2017unsupervised} can be extended to deal with new domains that may be encountered after the initial training.
In the computer vision literature, several adaptations of batch normalization (BN)~\cite{ioffe2015batch} have been suggested for domain adaptation~\cite{li2016revisiting,carlucci2017autodial} and multi-domain learning~\cite{bilen2017universal,rebuffi2017learning} for object recognition using CNNs.
Broadly, these works employ BN for domain-specific scaling to account for domain shifts, while sharing the bulk of the CNN parameters to leverage the similarity between the domains.

In this work, we extend approaches based on adaptive BN layers for segmentation across scanning protocols in a lifelong learning setting.
In particular, we train a CNN with common convolutional filters and specific BN parameters for each protocol/scanner.
The network is initially trained with images from a few scanners to learn appropriate convolutional filters.
By fine-tuning the BN parameters with a few labelled images, it can then be adapted to new protocols/scanners.
Crucially, this is achieved without performance degradation on the older scanners, whose training data is not available after the initial training.
\section{Method}\label{sec:methods}
Batch normalization (BN) was introduced in~\cite{ioffe2015batch} to enable faster training of deep neural networks by preventing saturated gradients via normalization of inputs before each non-linear activation layer.
In a BN layer, each batch $x_B$ is normalized as shown in Eq.~\ref{BN_equation}.
During training, $\mu_B$ and $\sigma^2_B$ are the mean and variance of $x_B$,
while at test time, they are the estimated population mean and variance as approximated by a moving average over training batches.
$\gamma$, $\beta$ are learnable parameters that allow the network to undo the normalization, if required.
Inspired by~\cite{bilen2017universal}, we propose to use separate batch normalization for each protocol / scanner.

\begin{equation}~\label{BN_equation}
BN(x_B) = \gamma\times\dfrac{x_B - \mu_B}{\sqrt{\sigma^2_B+\epsilon}} + \beta
\end{equation}

Notwithstanding variations in image statistics due to inter-scanner differences, a segmentation network would be confronted with images of the same organ, acquired with the same modality (MR).
Hence, it is reasonable to postulate common characteristics between the domains and thus, shared support in an appropriate representation space.
Following~\cite{bilen2017universal}, we hypothesize that such a representation space can be found by using domain-agnostic convolutional filters and that the inter-domain differences can be handled by appropriate normalization via domain-specific BN modules.
This approach is not only in line with the previous domain adaptation works~\cite{carlucci2017autodial}, but also embodies the normalization idea of conventional proposals for dealing with inter-scanner variations~\cite{zhuge2009intensity,weisenfeld2004normalization,han2007atlas}.
Further, like~\cite{rebuffi2017learning}, once suitable shared convolutional filters have been learned, we adapt the domain-specific BN layers to new related domains.

The training procedure in our framework is as follows.
We use superscript $^{bn}$ to indicate a network with domain-specific BN layers.
We initially train a network, $N_{12\cdots d}^{bn}$ on $d$ domains, with shared convolutional filters and separate BN parameters, $bn_{k}$, for each domain $D_k$.
During training, each batch consists of only one domain, with all domains covered successively.
In a training iteration when the batch consists of domain $D_k$, $bn_{k'}$ for k'$\neq$k are frozen. 
Now, consider a new domain $D_{d+1}$, with a few labelled images $I_{D_{d+1}}$.
We split this small dataset into two halves, using one for training, $I_{D_{d+1}}^{tr}$ and the other for validation, $I_{D_{d+1}}^{vl}$.
We evaluate the performance of $N_{12\ldots d}^{bn}$ on $I_{D_{d+1}}^{tr}$, using each learned $bn_{k}$, $k=1,2,\cdots d$.
If $bn_{k^*}$ leads to the best accuracy, we infer that among the already learned domains, $D_{k^*}$ is the closest to $D_{d+1}$.
Then, keeping the convolutional filter weights fixed, an additional set of BN parameters $bn_{d+1}$ is initialized with $bn_{k^*}$ and fine-tuned using $I_{D_{d+1}}^{tr}$ with standard stochastic gradient descent minimization.
The optimization is stopped when the performance on $I_{D_{d+1}}^{vl}$ stops improving.
Now, the network can segment all domains $D_k$, for $k=1,2,\ldots d, d+1$ using their respective $bn_k$.

In the spirit of lifelong learning, this approach allows learning on new domains with only a few labelled examples.
This is enabled by utilizing the knowledge obtained from learning on the old domains, in the form of the trained domain-agnostic parameters.
The fact that the number of domain-specific parameters is small comes with two advantages.
One, that they can be tuned for a new domain by training with a few labelled images quickly and with minimal risk of overfitting.
Secondly, they can be saved for each domain without significant memory footprint.
Finally, catastrophic forgetting~\cite{french1999catastrophic} by performance degradation on previous domains does not arise in this approach by construction because of the explicit separate modeling of shared and private parameters.


\section{Experiments and Results}
\textbf{Datasets:}
Brain MR datasets from several scanners, hospitals, or acquisition protocols are required to test the applicability of the proposed method for lifelong multi-domain learning.
To the best of our knowledge, there are only a few publicly available brain MRI datasets with ground truth segmentation labels from human experts.
Therefore, we use FreeSurfer~\cite{fischl2012freesurfer} to generate pseudo ground truth annotations.
While annotations from human experts would be ideal, we believe that FreeSurfer annotations can serve as a reasonable proxy to test our approach to lifelong multi-scanner learning.

We use images from 4 publicly available datasets: Human Connectome Project (HCP)~\cite{van2013wu}, Alzheimer’s Disease Neuroimaging Initiative (ADNI)\footnote{adni.loni.usc.edu}, Autism Brain Imaging Data Exchange (ABIDE)~\cite{di2014autism} and Information eXtraction from Images (IXI)\footnote{brain-development.org/ixi-dataset/}.
The datasets are split into different domains, as shown in Table~\ref{tab:dataset_details}.
Domains $D_1$, $D_2$, $D_3$ are treated as initially available, and $D_4$, $D_5$ as new.
The number of training and test images for each domain indicated in the table are explained later while describing the experiments.

\vspace{0.15cm} \noindent \textbf{Training details:} 
While the domain-specific BN layers can be incorporated in any standard CNN, we work with the widely used U-Net~\cite{ronneberger2015u} architecture with minor alterations.
Namely, our network has a reduced depth with three max-pooling layers and a reduced number of kernels: {32,64,128,256} in the convolutional blocks on the contracting path and {128,64,32} on the upscaling path.
Also, bilinear interpolation is preferred to deconvolutional layers for upscaling in view of potential checkerboard artifacts~\cite{odena2016deconvolution}.
The network is trained to minimize the dice loss, as introduced in~\cite{milletari2016v} to reduce sensitivity to imbalanced classes.
Per image volume, the intensities are normalized by dividing with their 98 \%tile.
The initial network trains in about 6 hours, while the domain-specific BN modules can be updated for a new domain in about 1 hour.

\begin{table}[t!]
\caption{Details of the datasets used for our experiments.}\label{tab:dataset_details}
\centering
\setlength{\tabcolsep}{4pt}
   \begin{tabular}{c|c|c|c|c|c|c}
		\hline
		Domain & Dataset & Field & MR Modality & $n_{train}$ & $n_{train}^{scratch}$ & $n_{test}$\\
		\hline
		D$_{1}$ & HCP & 3T & T1w & 30 & 30 & 20\\
        D$_{2}$ & HCP & 3T & T2w & 30 & 30 & 20\\
        D$_{3}$ & ADNI & 1.5T & T1w & 30 & 30 & 20\\
        D$_{4}$ & ABIDE, Caltech & 3T & T1w & 4 & 20 & 20\\
        D$_{5}$ & IXI & 3T & T2w & 4 & 20 & 20\\
	\end{tabular} \vspace{-0.5cm}
\end{table} 

\vspace{0.15cm} \noindent \textbf{Experiments:}
We train three types of networks, as described below.
\begin{itemize}
\item Individual networks $N_d$: Trained for each domain $d$, with $n_{train}^{scratch}$ training images (see Table~\ref{tab:dataset_details}).
For the known domains ($D_1$, $D_2$, $D_3$), the accuracy of $N_d$ serves as a baseline that the other networks with shared parameters must preserve.
For the new domains ($D_4$, $D_5$), the performance of $N_d$ is the benchmark that we seek to achieve by training on much fewer training examples ($n_{train}$) and using the knowledge of the previously learned domains.

\item A shared network $N_{123}$: Trained on $D_1$, $D_2$, $D_3$ with $n_{train}$ images, with all parameters shared including the BN layers, bn$_s$.
In contrast to the training regime of $N^{bn}_{1,2,\ldots d}$ described in Section~\ref{sec:methods}, while training $N_{123}$ each batch randomly contains images from all domains to ensure that the shared BN parameters can be tuned for all domains.
Histogram equalization~\cite{nyul2000new} is applied to a new domain $D_d$ before being tested $N_{123}$.
For adapting $N_{123}$ to $D_d$, its parameters are fine-tuned with $n_{train}$ images of the new domain and the modified network is referred to as $N_{123\to d}$.

\item A lifelong multi-domain learning network $N_{123}^{bn}$: Trained on $D_1$, $D_2$, $D_3$, with shared convolutional layers and domain-specific BN layers.
The updated network after extending $N_{123}^{bn}$ for a new domain $D_d$ according to the procedure described in Sec.~\ref{sec:methods} is called $N_{123,k^*\to d}$, where $k^*$ is the closest domain to $D_d$.
\end{itemize}

\begin{figure}[h!]
\setlength{\belowcaptionskip}{-20pt}
\centering
\includegraphics[trim = 45mm 23mm 41mm 15mm, clip, width=0.192\textwidth]{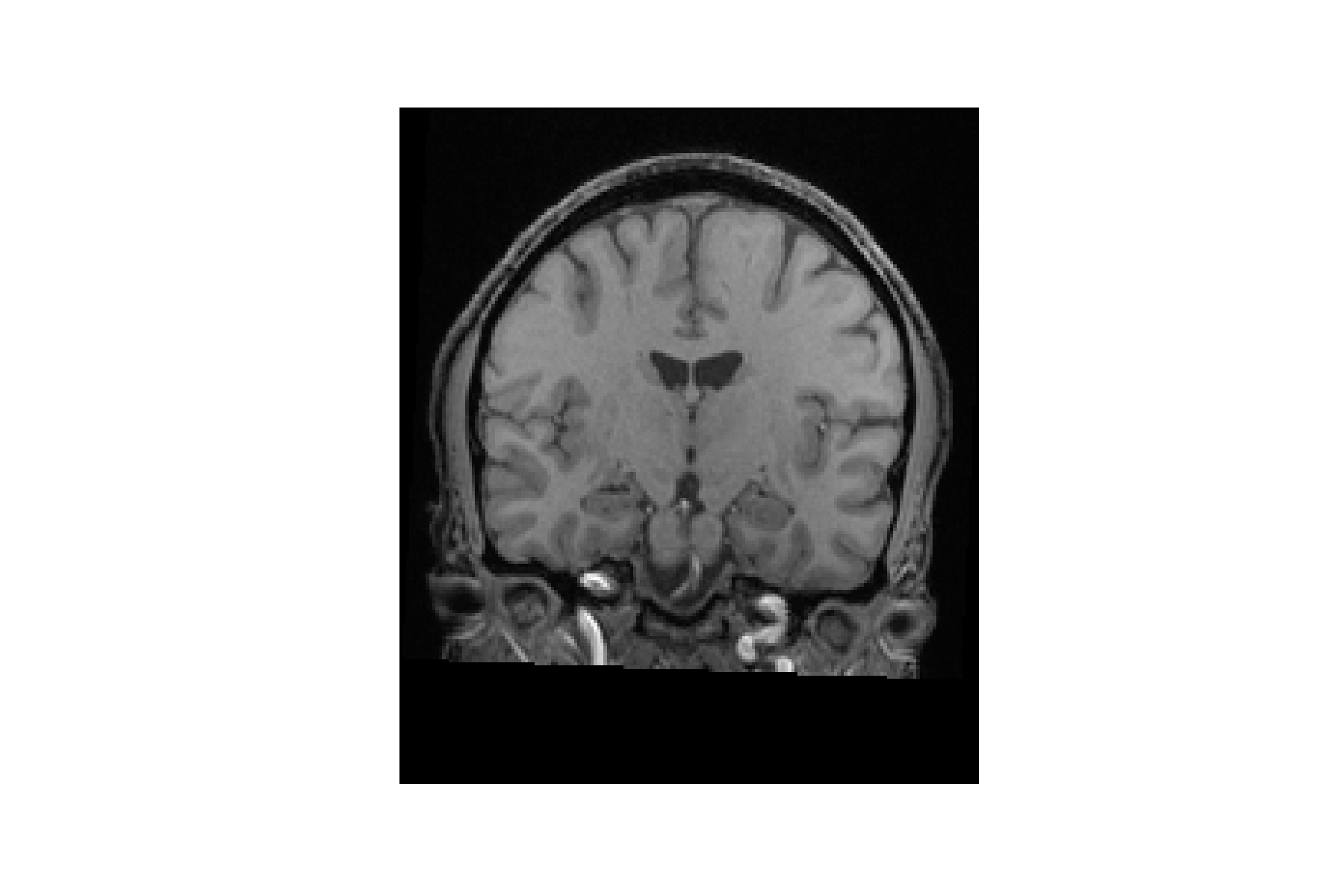}
\includegraphics[trim = 45mm 23mm 41mm 15mm, clip, width=0.192\textwidth]{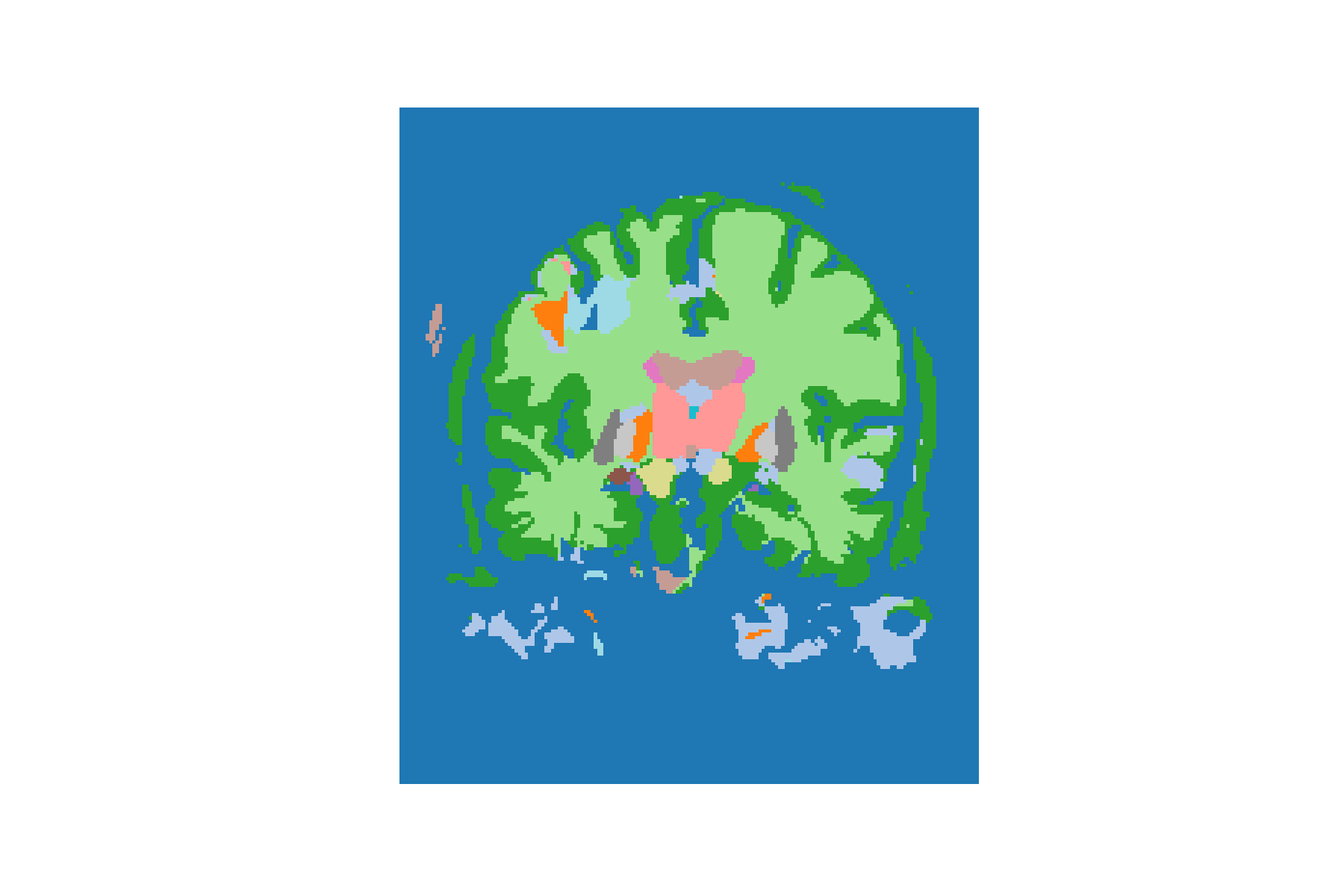}
\includegraphics[trim = 45mm 23mm 41mm 15mm, clip, width=0.192\textwidth]{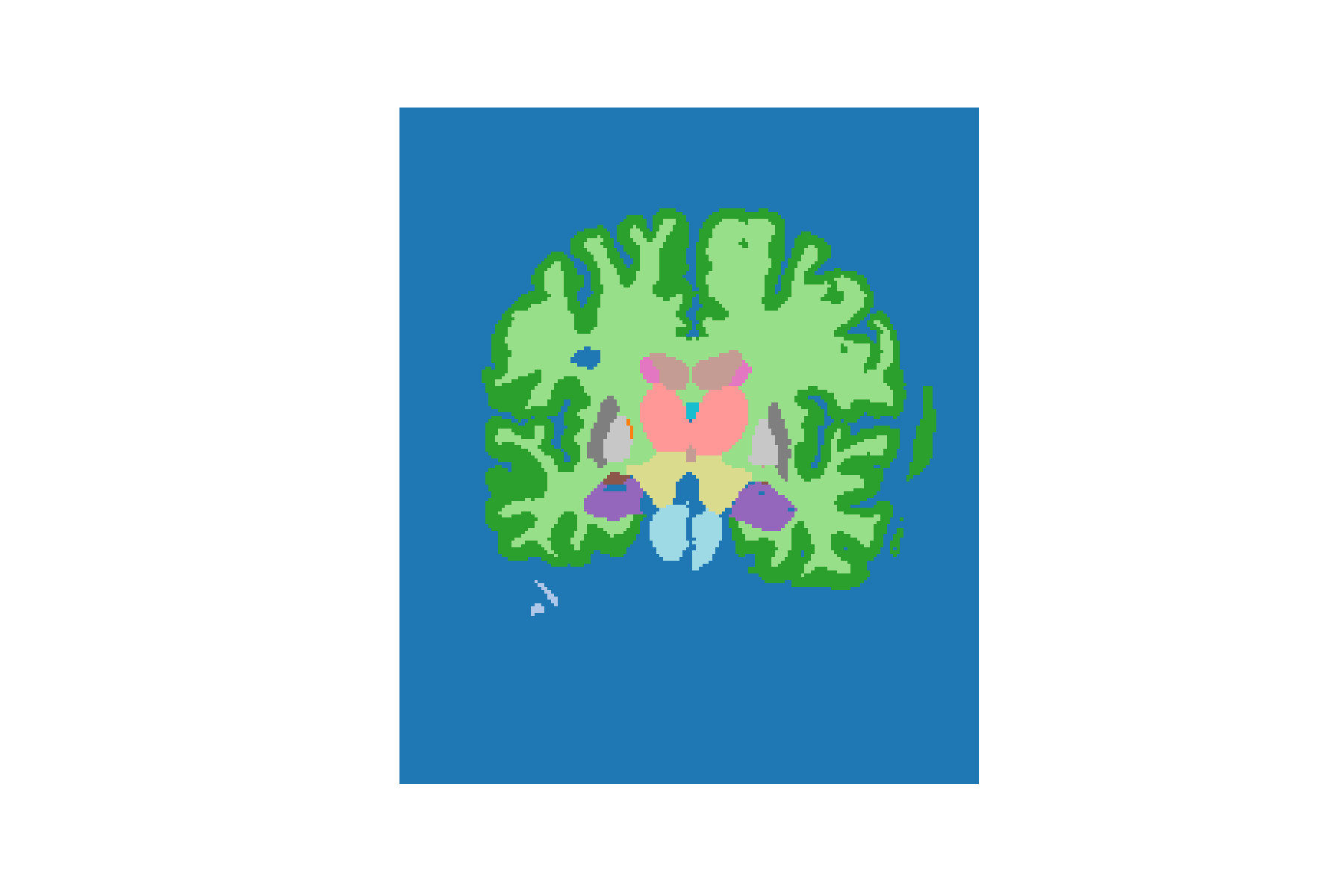}
\includegraphics[trim = 45mm 23mm 41mm 15mm, clip, width=0.192\textwidth]{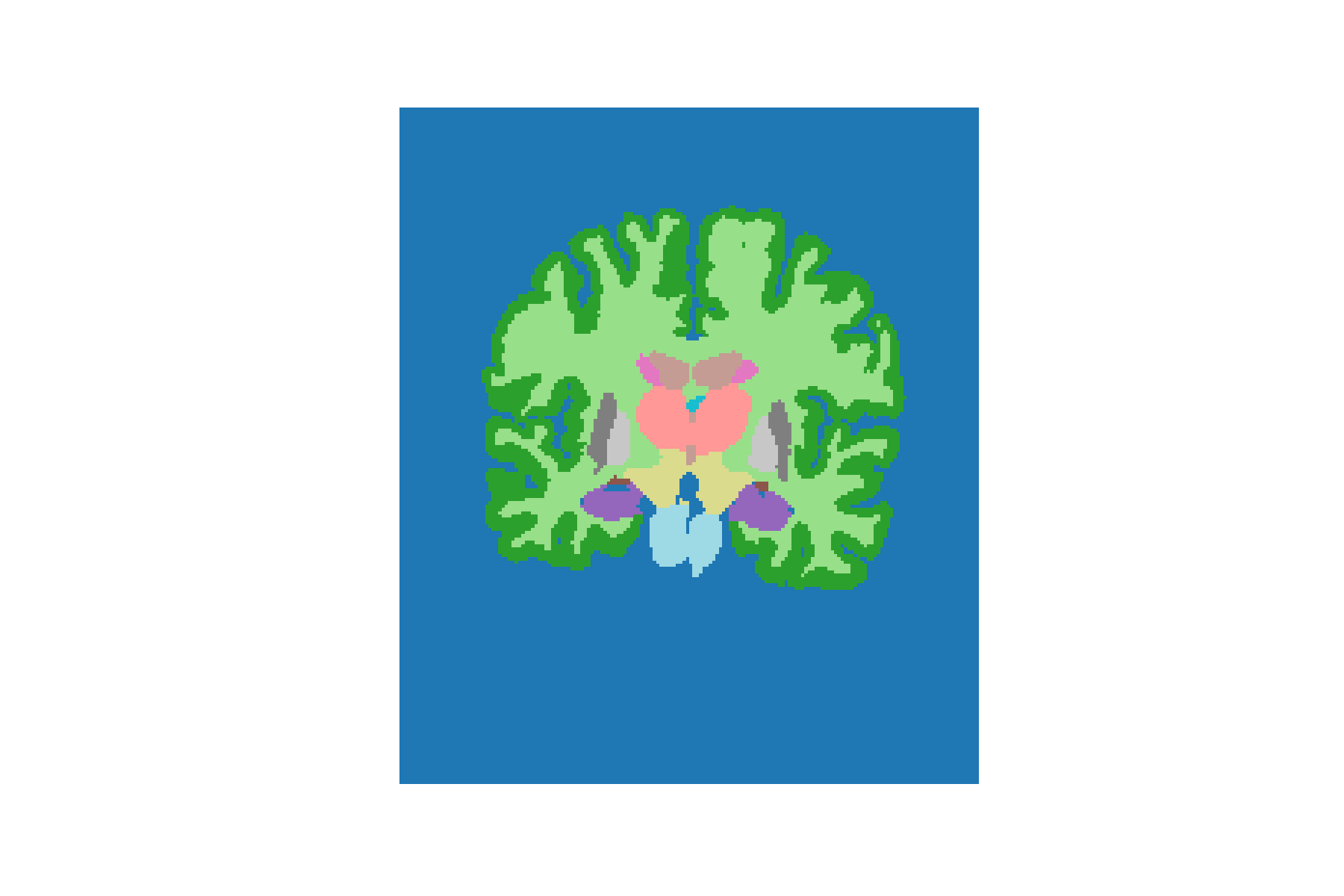}
\includegraphics[trim = 45mm 23mm 41mm 15mm, clip, width=0.192\textwidth]{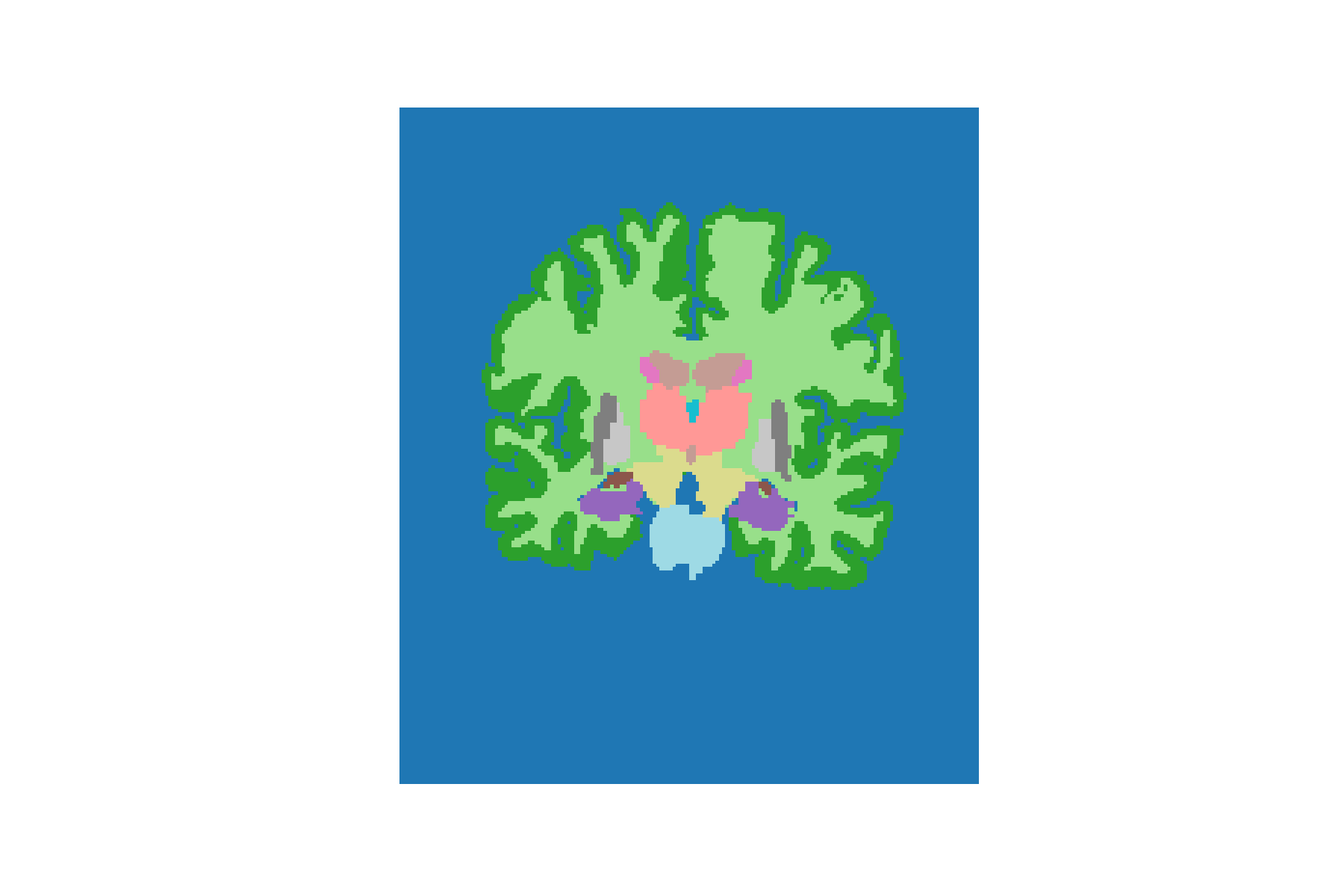}
\includegraphics[trim = 45mm 23mm 41mm 15mm, clip, width=0.192\textwidth]{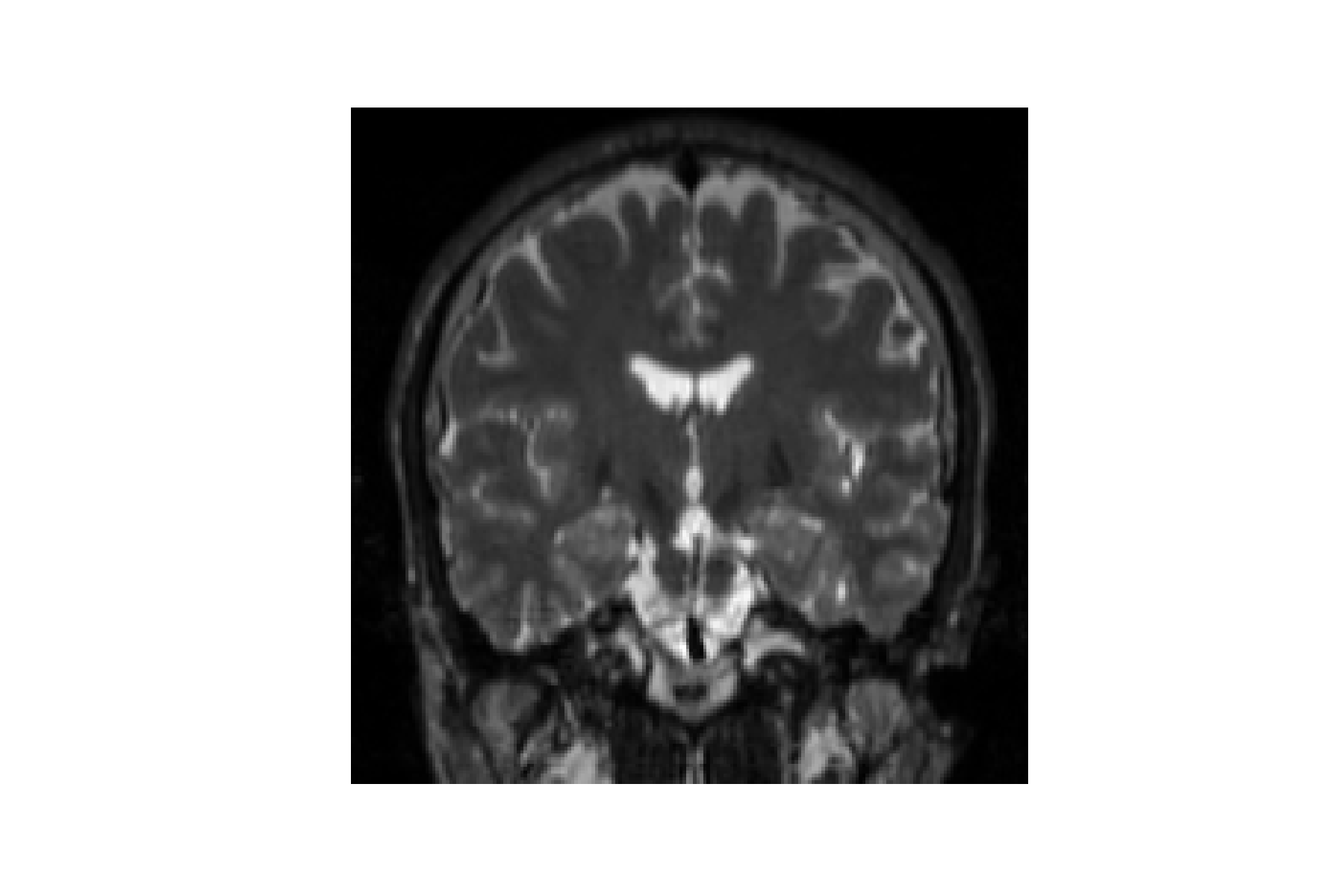}
\includegraphics[trim = 45mm 23mm 41mm 15mm, clip, width=0.192\textwidth]{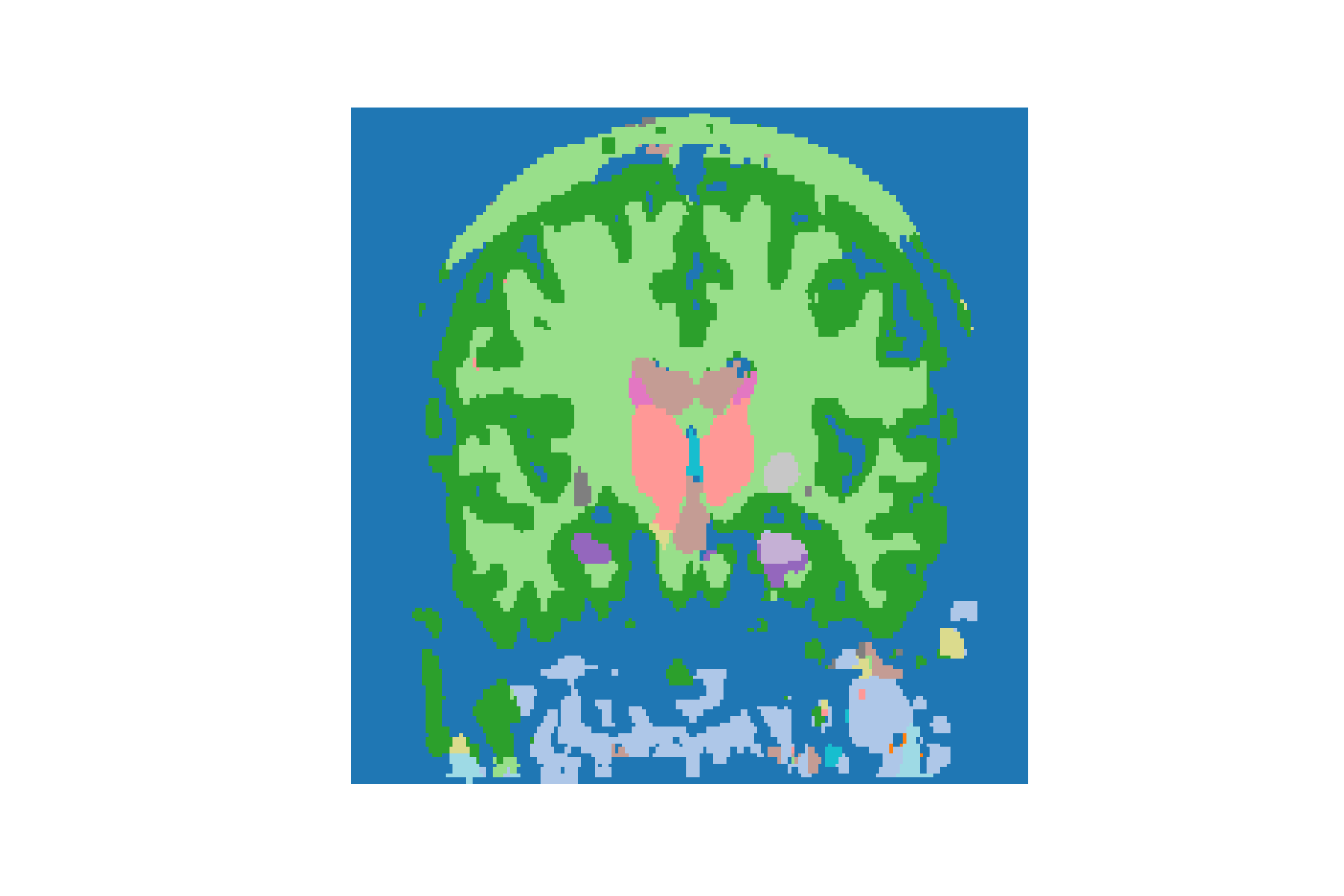}
\includegraphics[trim = 45mm 23mm 41mm 15mm, clip, width=0.192\textwidth]{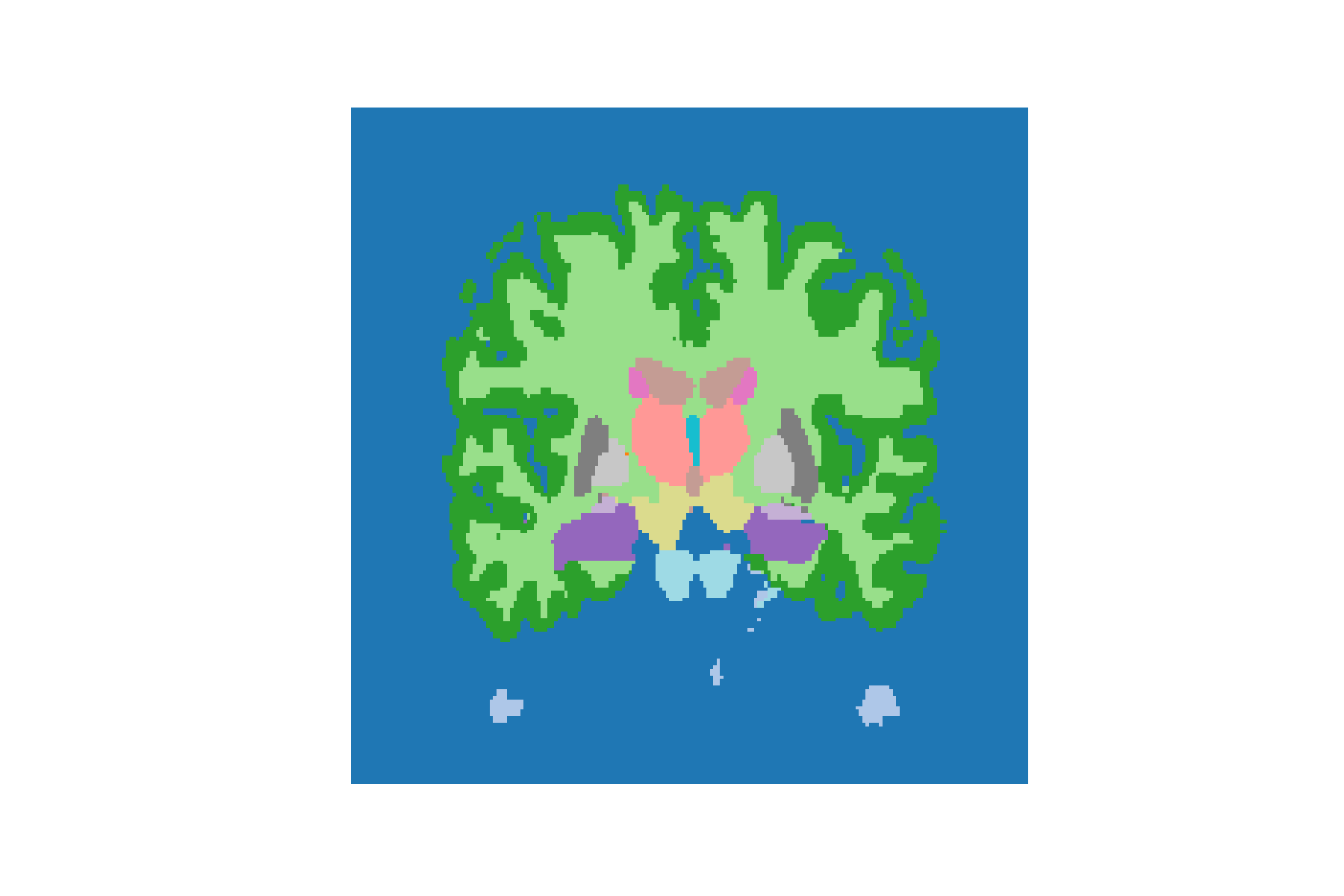}
\includegraphics[trim = 45mm 23mm 41mm 15mm, clip, width=0.192\textwidth]{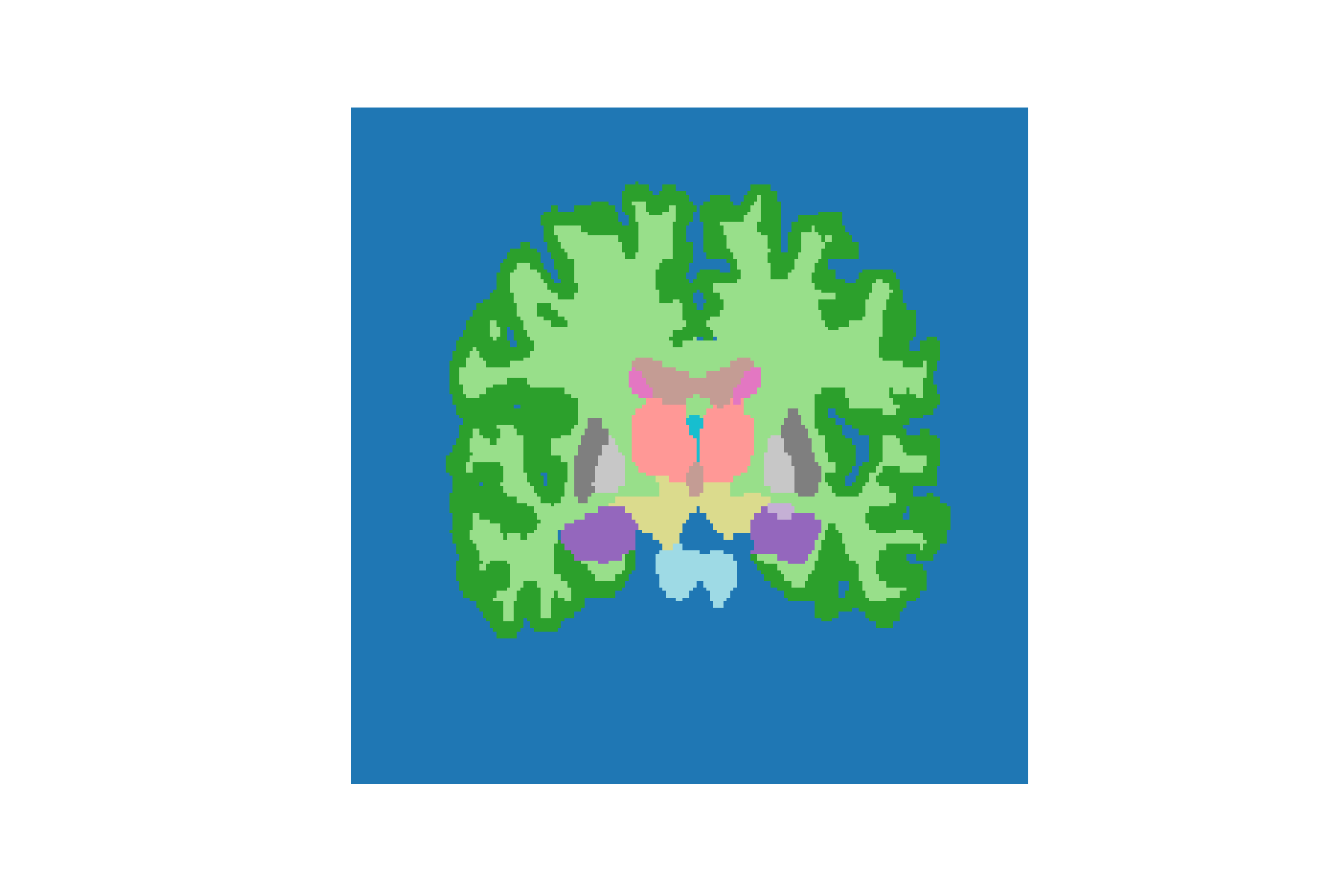}
\includegraphics[trim = 45mm 23mm 41mm 15mm, clip, width=0.192\textwidth]{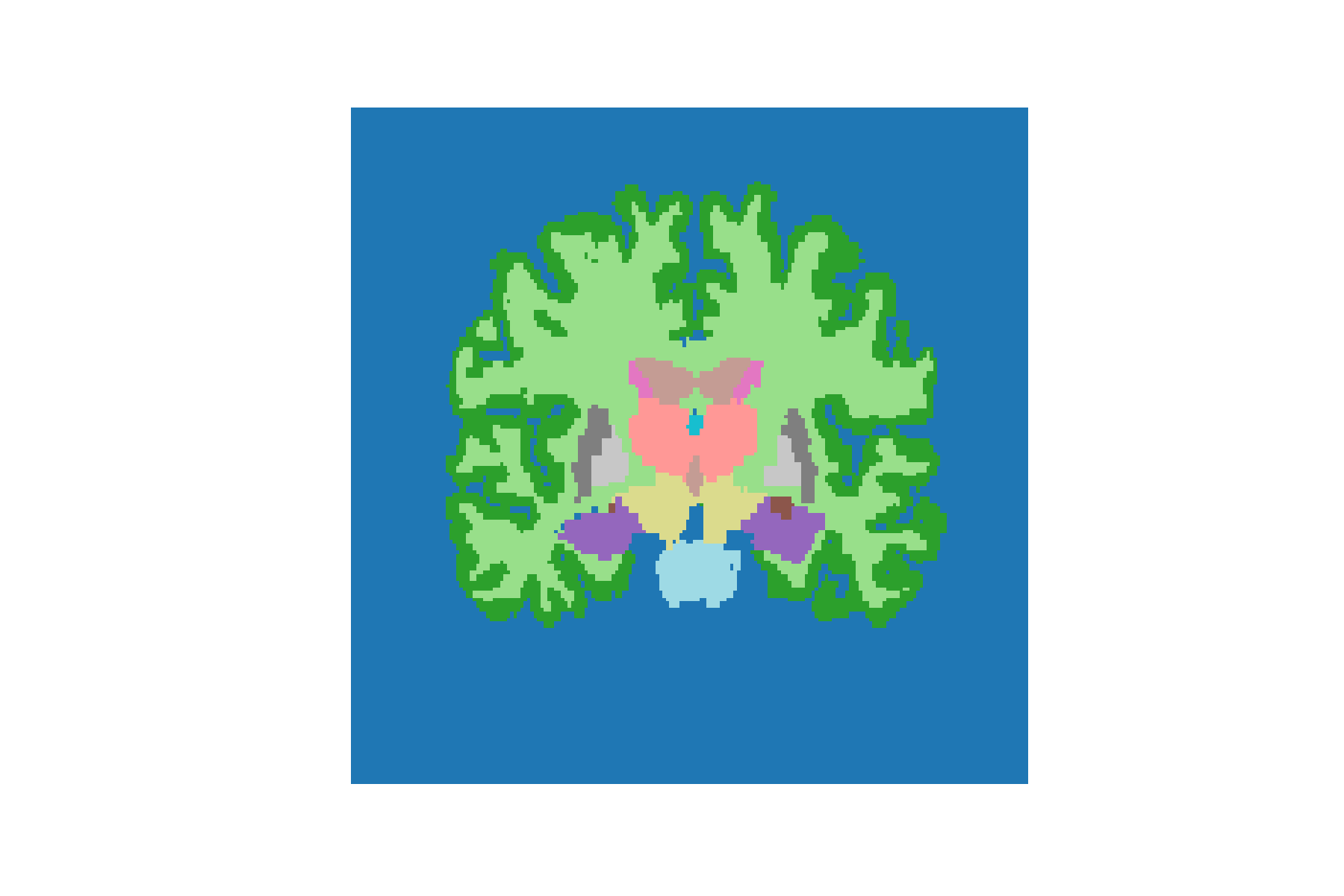}
\hspace{2.1cm} a \hspace{2.1cm} b \hspace{2.1cm} c \hspace{2.1cm} d \hspace{2.1cm} e
\caption{Qualitative results: (a) images from domains D$_d$, segmentations predicted by (b) N$_{123}^{bn}$, bn$_{k^*}$, (c) N$_{123,k^*\to d}^{bn}$, bn$_d$, (d) N$_{d}$ and (e) ground truth annotations, with \{$d$, $k^*$\} as \{4, 3\} (top) and \{5, 2\} (bottom).}
\label{fig:qualitative_results}
\end{figure}

\vspace{0.15cm} \noindent \textbf{Results:}
All networks are evaluated based on their mean Dice score for $n_{test}$ images from the appropriate domain (see Table~\ref{tab:dataset_details}).
Quantitative results of our experiments are shown in Table~\ref{tab:quant_results}.
The findings can be summarized as follows:
\begin{itemize}
\item $N_{123}$ preserves the performance of $N_{1}$, $N_{2}$, $N_{3}$.
Thus, a single network can learn to segment multiple domains, provided sufficient training data is available from all the domains at once.
However, its performance severely degrades for unseen domains $D_4$ and $D_5$.
Histogram equalization (denoted by $D_{d,HistEq}$) to the closest domain is unable to improve performance significantly, while fine-tuning the network for the new domains causes catastrophic forgetting~\cite{french1999catastrophic}, that is, degradation in performance on the old domains.

\item $N_{123}^{bn}$ also preserves the performance of $N_{1}$, $N_{2}$, $N_{3}$.
For a new domain $D_4$, using the $bn_3$ parameters of the trained $N_{123}^{bn}$ lead to the best performance.
Thus, we infer that $D_3$ is the closest to $D_4$ among $D_1$, $D_2$, $D_3$.
After fine-tuning the parameters of BN$_3$ to obtain those of BN$_4$, the dice scores for all the structures improve dramatically and are comparable to the performance of $N_4$.
Crucially, as the original $bn_k$ for k=1,2,3 are saved, the performance on $D_1$, $D_2$, $D_3$ in the updated network $N_{123,3-4}^{bn}$ is exactly the same as in $N_{123}^{bn}$.
Similar results can be seen for the other new domain, $D_5$.
The improvement in the segmentations for new domains after fine-tuning the BN parameters can also be observed qualitatively in Fig.~\ref{fig:qualitative_results}.

\end{itemize}

\begin{table}[t!]
\caption{Segmentation Dice scores for different domains for the three different types of networks, trained as explained in the experiments section.}\label{tab:quant_results}
\centering
\setlength{\tabcolsep}{4pt}
   \begin{tabular}{lll|ccccccc|c}
		\Xhline{2\arrayrulewidth}
		Network & Test & BN & Thal & Hipp & Amyg & Ventr & Caud & Puta & Pall & Avg\\
		\Xhline{2\arrayrulewidth}
		N$_{1}$ & D$_{1}$ & bn$_{1}$ & 0.919 & 0.861 & 0.849 & 0.901 & 0.9 & 0.887 & 0.747 & 0.866 \\
        N$_{2}$ & D$_{2}$ & bn$_{2}$ & 0.912 & 0.84 & 0.836 & 0.891 & 0.889 & 0.876 & 0.736 & 0.854 \\
        N$_{3}$ & D$_{3}$ & bn$_{3}$ & 0.913 & 0.872 & 0.81 & 0.944 & 0.864 & 0.879 & 0.853 & 0.876 \\
        N$_{4}$ & D$_{4}$ & bn$_{4}$ & 0.924 & 0.879 & 0.853 & 0.933 & 0.912 & 0.9 & 0.851 & 0.893 \\
        N$_{5}$ & D$_{5}$ & bn$_{5}$ & 0.884 & 0.79 & 0.773 & 0.803 & 0.793 & 0.818 & 0.791 & 0.81 \\
		\Xhline{3\arrayrulewidth}
        N$_{123}$ & D$_{1}$ & bn$_{s}$ & 0.909 & 0.846 & 0.824 & 0.891 & 0.878 & 0.877 & 0.745 & 0.853 \\
        N$_{123}$ & D$_{2}$ & bn$_{s}$ & 0.888 & 0.838 & 0.815 & 0.876 & 0.863 & 0.86 & 0.701 & 0.834 \\
        N$_{123}$ & D$_{3}$ & bn$_{s}$ & 0.905 & 0.851 & 0.792 & 0.938 & 0.863 & 0.873 & 0.828 & 0.864 \\
		\hline
        N$_{123}$ & D$_{4}$ & bn$_{s}$ & 0.745 & 0.249 & 0.057 & 0.787 & 0.428 & 0.324 & 0.071 & 0.38 \\
        N$_{123}$ & D$_{4,HistEq}$ & bn$_{s}$ & 0.641 & 0.428 & 0.175 & 0.754 & 0.628 & 0.579 & 0.303 & 0.501 \\
        N$_{123\to 4}$ & D$_{4}$ & bn$_{s}$ & 0.91 & 0.856 & 0.74 & 0.922 & 0.894 & 0.859 & 0.786 & 0.852 \\
        N$_{123\to 4}$ & D$_{1}$ & bn$_{s}$ & 0.869 & 0.809 & 0.773 & 0.867 & 0.861 & 0.722 & 0.667 & 0.795 \\
        N$_{123\to 4}$ & D$_{2}$ & bn$_{s}$ & 0.676 & 0.418 & 0.512 & 0.105 & 0.635 & 0.4 & 0.411 & 0.451 \\
        N$_{123\to 4}$ & D$_{3}$ & bn$_{s}$ & 0.801 & 0.762 & 0.65 & 0.753 & 0.728 & 0.715 & 0.772 & 0.74 \\
        \hline
        N$_{123}$ & D$_{5}$ & bn$_{s}$ & 0.418 & 0.178 & 0.182 & 0.438 & 0.268 & 0.197 & 0.025 & 0.244 \\
        N$_{123}$ & D$_{5,HistEq}$ & bn$_{s}$ & 0.294 & 0.143 & 0.16 & 0.437 & 0.261 & 0.293 & 0.01 & 0.228 \\
        N$_{123\to 5}$ & D$_{5}$ & bn$_{s}$ & 0.861 & 0.777 & 0.761 & 0.799 & 0.76 & 0.796 & 0.741 & 0.785 \\
        N$_{123\to 5}$ & D$_{1}$ & bn$_{s}$ & 0.267 & 0.022 & 0.173 & 0.004 & 0.05 & 0.002 & 0.004 & 0.075 \\
        N$_{123\to 5}$ & D$_{2}$ & bn$_{s}$ & 0.574 & 0.574 & 0.564 & 0.739 & 0.657 & 0.521 & 0.526 & 0.594 \\
        N$_{123\to 5}$ & D$_{3}$ & bn$_{s}$ & 0.147 & 0.029 & 0.16 & 0.006 & 0.114 & 0.039 & 0.003 & 0.071 \\
        \Xhline{3\arrayrulewidth}
        N$_{123}^{bn}$ & D$_{1}$ & bn$_{1}$ & 0.916 & 0.852 & 0.84 & 0.894 & 0.893 & 0.884 & 0.729 & 0.858 \\
        N$_{123}^{bn}$ & D$_{2}$ & bn$_{2}$ & 0.91 & 0.853 & 0.843 & 0.887 & 0.882 & 0.873 & 0.749 & 0.857 \\
        N$_{123}^{bn}$ & D$_{3}$ & bn$_{3}$ & 0.911 & 0.868 & 0.818 & 0.944 & 0.867 & 0.879 & 0.846 & 0.876 \\
        \hline
        N$_{123}^{bn}$ & D$_{4}$ & bn$_{1}$ & 0.621 & 0.288 & 0.218 & 0.173 & 0.676 & 0.576 & 0.457 & 0.43 \\
        N$_{123}^{bn}$ & D$_{4}$ & bn$_{2}$ & 0.162 & 0 & 0.001 & 0.001 & 0.04 & 0.017 & 0 & 0.032 \\
        N$_{123}^{bn}$ & D$_{4}$ & bn$_{3}$ & 0.721 & 0.271 & 0.305 & 0.549 & 0.569 & 0.515 & 0.297 & 0.461 \\
        N$_{123,3\to 4}^{bn}$ & D$_{4}$ & bn$_{4}$ & 0.878 & 0.83 & 0.772 & 0.907 & 0.875 & 0.852 & 0.772 & 0.841 \\
        \hline
        N$_{123}^{bn}$ & D$_{5}$ & bn$_{1}$ & 0.001 & 0.019 & 0.062 & 0.008 & 0.004 & 0 & 0 & 0.013 \\
        N$_{123}^{bn}$ & D$_{5}$ & bn$_{2}$ & 0.354 & 0.123 & 0.268 & 0.225 & 0.407 & 0.276 & 0.366 & 0.288 \\
        N$_{123}^{bn}$ & D$_{5}$ & bn$_{3}$ & 0 & 0.003 & 0.031 & 0.001 & 0 & 0 & 0 & 0.005 \\
        N$_{123,2\to 5}^{bn}$ & D$_{5}$ & bn$_{5}$ & 0.774 & 0.687 & 0.687 & 0.761 & 0.669 & 0.714 & 0.713 & 0.715 \\
		\Xhline{3\arrayrulewidth}
	\end{tabular}
\end{table}

\section{Conclusion}
In this article, we presented a lifelong multi-domain learning approach to learn a segmentation CNN that can be for related MR modalities and across scanners/protocols.
Further, it can be adapted to new scanners or protocols with only a few labelled images and without degrading performance on the previous scanners.
This was achieved by learning batch normalization parameters for each scanner, while sharing the convolutional filters between all scanners.
In future work, we intend to investigate the possibility of extending this approach to MR modalities that were not present during the initial training.

To the best of our knowledge, this is the first work to tackle the lifelong machine learning problem for CNNs in the context of medical image analysis.
We believe that this may set an important precedent for more research in this vein to handle data distribution changes which are ubiquitous in clinical data.


\bibliographystyle{IEEEbib}
\bibliography{main}





\end{document}